\documentclass[letterpaper, 10 pt, conference]{ieeeconf}  

\IEEEoverridecommandlockouts                              

\usepackage{times}
\usepackage{multicol}
\usepackage{latexsym}
\usepackage{color}
\usepackage{amsmath}
\usepackage{cite}     
\usepackage{listings}
\usepackage{mathrsfs}
\usepackage{amssymb}
\usepackage{pstool}
\usepackage{psfrag}
\usepackage[ruled]{algorithm2e}
\usepackage{hyperref}
\usepackage[table]{xcolor}
\usepackage{graphicx}
\usepackage{amsthm}
\usepackage{color,soul}
\usepackage{xcolor}
\usepackage{lineno,hyperref}
\usepackage{epsfig}
\usepackage{subfigure}
\usepackage{amssymb}
\usepackage{siunitx}
\usepackage{latexsym}
\usepackage{amsmath}
\usepackage{amsfonts}
\usepackage{blindtext}
\usepackage{scrextend}
\usepackage{pbox}
\usepackage{xcolor}
\usepackage{svg}
\usepackage[most]{tcolorbox}
\usepackage{amsthm}
\usepackage{float}

\newtheorem{assumption}{Assumption}

\title{\LARGE \bf
Trust Through Transparency: Explainable Social Navigation for Autonomous Mobile Robots via Vision-Language Models}

\author{Oluwadamilola Sotomi$^{1}$, Devika Kodi$^{1}$, and Aliasghar Arab$^{* 1, 2}$
\thanks{This work was supported by the department of Mechanical and Aerospace Engineering and New York University, Tandon School of Engineering}
\thanks{$^{1}$ Aliasghar and other authors are with the department of Mechanical and Aerospace Engineering, Tandon School of Engineering, New York University, 6 Metro Tech, Brooklyn, NY, USA
        {\tt\small aliasghar.arab@nyu.edu.}}%
\thanks{$^{2}$Aliasghar is with GenAuto.ai by General Autonomy Inc., 201 Centennial Ave., Piscataway, Nj, USA.
        {\tt\small mojarab@genauto.ai.}}%
}
\begin{document}
\maketitle
\thispagestyle{empty}
\pagestyle{empty}

\begin{abstract}
Service and assistive robots are increasingly being deployed in dynamic social environments; however, ensuring transparent and explainable interactions remains a significant challenge. This paper presents a multimodal explainability module that integrates vision language models and heat maps to improve transparency during navigation. The proposed system enables robots to perceive, analyze, and articulate their observations through natural language summaries. User studies (n=30) showed a preference of majority for real-time explanations, indicating improved trust and understanding. Our experiments were validated through confusion matrix analysis to assess the level of agreement with human expectations. Our experimental and simulation results emphasize the effectiveness of explainability in autonomous navigation, enhancing trust and interpretability.
\end{abstract}

\section{INTRODUCTION}
\color{black}{
As Autonomous Mobile Robots (AMRs) become increasingly integrated into social and service environments, ensuring safe and efficient navigation while interacting with humans remains a significant challenge~\cite{baraglia2017efficient}. Traditional AMRs often struggle to communicate their decision-making processes, leading to a lack of trust and usability in human-robot collaboration \cite{shekar2024explainable}. A fundamental requirement in Human Robot Interaction (HRI) is explainability. Robots must not only make decisions, but also communicate their reasoning in an intuitive manner to improve predictability and user confidence.
Transparency in robotic decision making fosters trust by helping users anticipate robot behavior and interact naturally\cite{lee2004trust}. Without it, humans struggle to adapt, leading to inefficiencies and hesitation. Although existing research has explored socially aware navigation models and explainable AI (XAI) in robotics, many approaches remain limited to internal decision logic, lacking human-readable real-time explanations~\cite{laban2023opening}. Furthermore, current systems often fail to incorporate multimodal reasoning, such as combining visual perception with language-based justifications~\cite{sanneman2022situation}.

\begin{figure}[ht!]
    \centering
    \includegraphics[width=.9\linewidth]{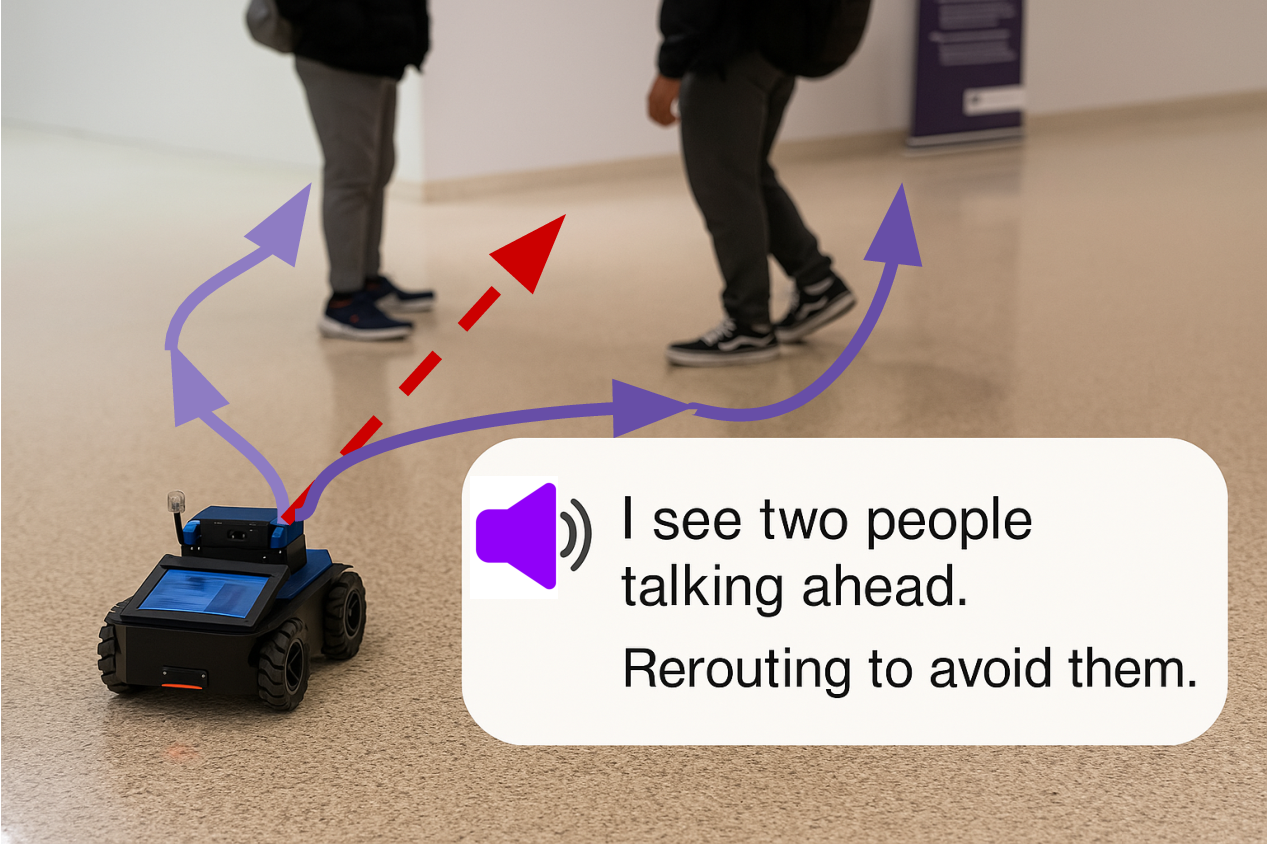}
    \caption{AMR approaches a social setting, demonstrating real-time explainable re-planning to avoid interrupting human interaction.}
    \label{fig:social}
\end{figure}

XAI plays a crucial role in improving human trust in autonomous systems. Early approaches used language models and prompt engineering for robot justifications, but lacked visual context, making explanations less intuitive. 
Recent studies incorporate Vision-Language Models (VLMs) to generate context-aware explanations by using cameras onboard~\cite{sobrin2024enhancing}. Explainability has also been explored in robot fault recovery, where natural language justifications assist users in diagnosing errors~\cite{das2021explainable}. Surrogate models, such as those based on Shapley values, improve decision transparency ~\cite{gavriilidis2023surrogate}. In addition, reinforcement learning (RL) approaches have used causal justifications based on Markov Decision Process (MDP) to improve policy interpretability ~\cite{finkelstein2022explainable}. These approaches highlight the importance of interpretable AI in improving human trust and usability in robotics~\cite{singh2024towards}~\cite{cruz2022evaluating} further evaluate how explanations in reinforcement learning scenarios align with human expectations, emphasizing the need for human-like justifications in real-world HRI settings. Parallelly, recent systems explore the use of vision-language models to improve HRI by allowing robots to understand and respond through more natural multimodal communication~\cite{abbas2024talkwithmachines}.

Social navigation requires robots to follow human norms. Traditional models like the Social Force Model (SFM) simulate human navigation but lack adaptability. Learning from Demonstration (LfD) has enabled robots to replicate human behaviors, though without high-level reasoning, leading to brittle responses. Recent efforts integrate language-based reasoning, encouraging datasets for perception, planning, and social navigation~\cite{payandeh2024social}. Risk-aware motion planning with multi-modal perception enhances safety in crowded environments. One method integrates Teb (Timed Elastic Band) with ORCA (Optimal Reciprocal Collision Avoidance) to refine real-time obstacle avoidance~\cite{wang2023motion}. Local path optimization using DWA and TEB planners in ROS improves narrow passage navigation and social compliance~\cite{yuan2022comparison}. However, beyond motion planning, robots must also integrate social reasoning for human-aware navigation.
Recent work integrates vision-language models with robot navigation, enabling socially aware behavior by scoring navigation decisions based on social norms and visual context~\cite{song2024vlm}. 

VLMs advance perception by enhancing situational awareness through text and visual data processing. Grad-CAM aids in interpretability by highlighting the salient image regions that influence robot decisions~\cite{selvaraju2020grad}. This improves trustworthiness in robotic applications by providing visual justifications. VLMs have also been explored for zero-shot semantic navigation, where they map visual input to frontier spaces for high-level planning without requiring task-specific training, as demonstrated in VLFM~\cite{yokoyama2024vlfm}. Beyond processing visual data, VLMs improve contextual understanding. BLIP (Bootstrapping Language-Image Pretraining) strengthens image-text grounding, allowing robots to generate context-aware descriptions~\cite{li2022blip}. This improves HRI, instruction following, and autonomous decision-making. Ensuring safe and explainable navigation remains a challenge. An AI-based assurance framework integrates XAI and security monitoring for real-time anomaly detection, enhancing safety and explainability in AI-driven autonomous systems~\cite{hamilton2020autonomous}.

To address these limitations, we introduce a multimodal explainability module that enables an AMR to generate human-readable, real-time explanations for its navigation behavior. Our approach leverages Vision-Language Foundation Models (VLFMs), integrating camera-based perception, heatmaps, and language models to articulate decisions. The cornerstone of our exploration lies in recognizing context-aware behavior and the explainability of AMRs around people to improve social acceptance. As new members of society, robots must take initiatives to be accepted by existing communities for future efficient contributions. The technological and social challenges of partially unknown interactions between robots and individuals have been studied, highlighting the disparities in the operational patterns that shape the robot environment. As illustrated in Fig.~\ref{fig:social}, the robot provides contextual explanations in natural language alongside heatmap-based visual reasoning, ensuring greater transparency in interactions.

Building on our previous work on explainability for robotic vehicles, this research extends our framework to AMRs by presenting more extensive experimental results and incorporating user surveys~\cite{shekar2024explainable}. We develop a ROS2-based explainability module that integrates a camera node, visual captioning using BLIP, Grad-CAM heatmaps for visual interpretability, and LLM-based natural language generation for real-time explanations. The interpretability of the framework is evaluated by measuring the accuracy of the explanation and alignment with human expectations through quantitative metrics. Furthermore, we demonstrate how integrating vision-language models with robotic navigation stacks enhances decision transparency and builds trust in human-robot interaction. Special attention is given to optimizing latency and ensuring real-time performance in dynamic environments. 

The remainder of this paper is structured as follows. Section II reviews related work, Section III details our methodology, Section IV presents experimental validation, and Section V concludes with future directions.

\begin{figure*}[ht!]
    \centering
    \includegraphics[width=0.9\linewidth]{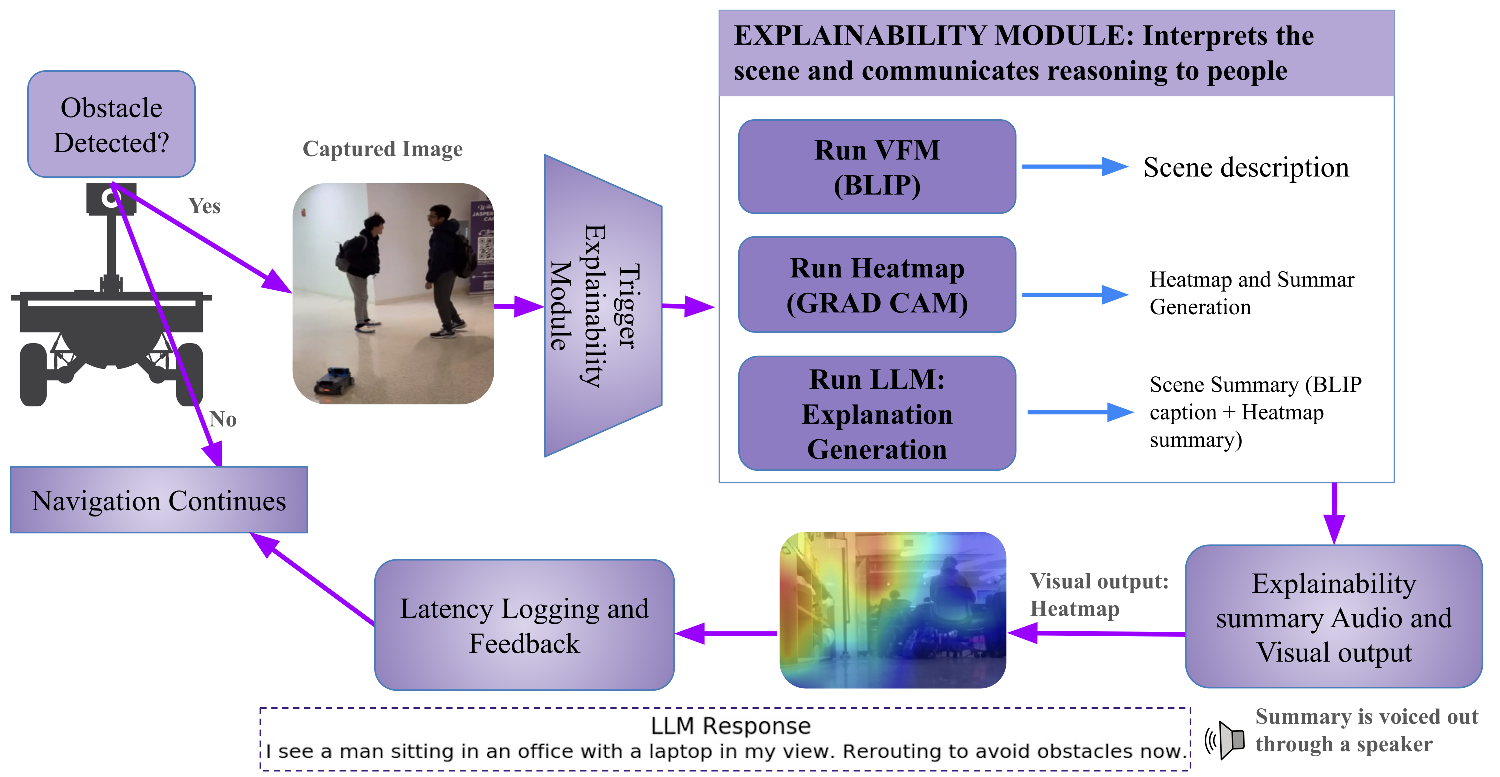}
    \caption{A diagram showing the relationship between the nodes that make up the explainability module.}
    \label{Node relationship}
\end{figure*}

\section{PROBLEM FORMULATION}
Autonomous mobile robotic systems operating in human-centered environments must adhere to predefined social norms to ensure safe and socially acceptable interactions by avoiding unnecessary navigation conflicts through explainability. We define the explainable mobile robot navigation task as a tuple
\begin{equation}
\label{eq:tuple}
\mathcal{T}_{\text{nav}} = \left( \mathcal{S}, \mathcal{G}, \mathcal{P}, \mathcal{E}, \varepsilon \right)
\end{equation}

\noindent where, \( \mathcal{S} = (\mathbf{q}, \mathbf{v}, \mathbf{q}_{\text{human}}) \) is the state of the robot, with \( \mathbf{q} \in \mathbb{R}^n \) as the position and orientation of the robot, \( \mathbf{v} \in \mathbb{R}^n \) as the velocity of the robot, \( \mathbf{q}_{\text{human}}^j \in \mathbb{R}^n \) as the observed position and orientation of the human $j^{\text{th}}$ from the robot's point of view. \( \mathcal{G} \equiv \mathbf{q}_g \in \mathbb{R}^n \) is the target configuration in the robot workspace. \( \mathcal{P} = \pi: [0, T] \rightarrow \mathbb{R}^n \) is the planned trajectory that maps time to robot location and velocities, so that the robot safely transitions from the initial state \( \mathbf{q}_0 \) to \( \mathbf{q}_g \) while avoiding obstacles and social conflicts with humans. \( \mathcal{E} = \{ e_t \mid t \in [0, T] \} \) is the set of multimodal explanations generated during execution, where each \( e_t \) includes interpretable outputs, such as descriptions of natural language through combination of visual heat maps, conditioned on the robot’s observations and decisions at time \( t \). \( \varepsilon \in [0, 1] \) is the explainability score reflecting the degree to which the system's behavior is interpretable to human observers, measured via user feedback or agreement metrics (e.g., confusion matrix alignment with human expectations).

The set of social constraints, human-centric safety requirements, and interaction rules can be formalized as a set of norm constraints \( \Omega_{norm} \), which must be satisfied at all times.

\begin{equation}
\Omega_{norm} = \bigcap_{i \in M} \Omega_i
\end{equation}

\noindent where \( \Omega_i \) represents the constraints imposed by the social norm \( i \) from the set of governing rules \( M \). For this purpose, we model these constraints in three different categories as suggested in~\cite{arab2021safe}
\begin{enumerate}
    \item \textbf{Human Safety and Social Norms:} The robot must maintain a safe distance from humans and adapt its trajectory to avoid discomfort as $\Omega_1$.
    \begin{equation}
    d_{\text{human}} \geq \max\{d_{\text{social}}, d_{\text{safe}}\},
    \end{equation}
\noindent where, $d_{\text{human}}$ is distance from the robot to human and $d_{\text{social}}$ and $d_{\text{safe}}$ are the safe and socially acceptable distance constants.
    \item \textbf{Socially Acceptable Motion:} The robot should avoid abrupt stops, excessive speed variations, or intrusive behaviors that could cause discomfort in human interactions, unless an aggressive maneuver is necessary to avoid an accident as $\Omega_2$.
    \begin{equation}
|\dot{\mathbf{v}}| \leq 
\begin{cases}
\text{No constraint} & \text{if } d_{\text{human}} \geq d_{\text{social}} \\
\alpha_{\text{social}} & \text{if } d_{\text{social}} \geq d_{\text{human}} \geq d_{\text{safe}} \\
\text{No constraint} & \text{if } d_{\text{human}} < d_{\text{safe}}
\end{cases}
\end{equation}
\noindent where, \( \mathbf{q} = [x, y, \psi] \) represent the robot pose in the odometry frame and \( \mathbf{v} = [v_x, v_y, \dot{\psi}] \) denote its velocity in the local frame. $\alpha_{\text{social}}$ is the maximum acceleration accepted in a social scenario for the robot. 
    \item \textbf{Social Navigation Constraints:} The robot should respect human space and avoid disrupting groups or ongoing interactions.
\begin{equation}
\label{eq:conflict}
h(\mathcal{P}, \mathbf{q}_{\text{human}}^{j}) \geq 0, \quad \forall \mathbf{q}_{\text{human}}^{j} \in \mathcal{M}_{\text{social}}
    \end{equation}
\end{enumerate}
\noindent where, $\mathcal{M}_{\text{social}}$ is the set of socially relevant human configurations (e.g., people conversing) and $h(\cdot) \geq 0$ encodes a social compliance or safety constraint. Any social conflict or non-safe situation should be represented by $h(\mathcal{P}, \mathbf{q}_{\text{human}}^{j}) \leq 0$ to ensure that no conflict occurs by satisfying Eq.~(\ref{eq:conflict}). By integrating socially aware constraints into navigation parameters, the proposed framework ensures that robot behavior remains predictable, interpretable, and aligned with human expectations, thus enhancing explainability and thus acceptability HRI.

\section{METHODOLOGY}
The objective is to calculate a safe, feasible and interpretable path \( P \), while maximizing \( \varepsilon \) through novel explainability modules, to improve transparency and trust during robot navigation in dynamic environments populated by humans. Our approach consists of three parts.
\begin{itemize}
    \item \textbf{1)} Development of a multimodal explainability system.
    \item \textbf{2)} Deployment in an AMRs for real-time validation.
    \item \textbf{3)} Integration with an autonomous navigation stack.
\end{itemize}

\begin{assumption}
The effectiveness of the explainability module is quantified by a scalar \emph{explainability factor} \( \varepsilon \in [0, 1] \), which reflects how well the robot's behavior is understood by users. The value of \( \varepsilon \) is determined through user feedback collected after the experiment via structured surveys that assess the clarity of the explanation, the alignment with human expectations, and the overall interpretability.

\begin{equation}
\varepsilon =
\begin{cases}
0, & \text{if explainability is inactive}, \\
\hat{\varepsilon} \in (0, 1], & \text{if explainability is active}.
\end{cases}
\end{equation}
\noindent where, \( \hat{\varepsilon} \) is a normalized score derived from survey responses and subjective evaluation metrics.
\end{assumption}

\begin{algorithm}[h!]
\caption{Explainablity Module via VLM, Heatmap and LLM}
\label{alg:ExplainableNavigation}
\nl Initialize LLM Node and Explainability Module\;
\nl Subscribe to topics \texttt{'camera/image'}, \texttt{'blip/caption'}, and \texttt{'heatmap/summary'}\;
\nl Set explainability factor \( \varepsilon \leftarrow 0 \)\;
\nl Set ExplainabilityModuleEnabled flag\;

\While{robot is navigating}{
    \nl Receive image from camera stream\;
    \nl Detect potential social conflict using VLM Node\;
    \nl Generate visual saliency map using Heatmap Node\;
    \If{conflict is detected}{
        \If{ExplainabilityModuleEnabled}{
            \nl Generate natural language explanation using LLM Node\;
            \nl Synthesize and output speech from explanation\;
            \nl Overlay and display heatmap with textual explanation\;
            \nl Save image, heatmap, and explanation with timestamp\;
            \nl Update explainability factor \( \varepsilon \leftarrow \varepsilon + \Delta\varepsilon \)\;}
        \nl Update navigation path to avoid conflict\;}
    \nl Execute current navigation step\;}
\nl Analyze navigation performance metrics (e.g., path efficiency, social acceptance)\;
\nl Correlate performance with explainability factor \( \varepsilon \)\;
\end{algorithm}

\begin{tcolorbox}[colback=gray!10!white, colframe=black, title=LLM Guiding Prompt, sharp corners=south, boxrule=0.5mm]
\textit{
"You are a mobile robot trying to avoid obstacles to reach your destination. The image caption is: '\{caption\}'. The heatmap analysis shows: '\{heatmap\_summary\}'. Provide a short, one-sentence description of your view. Do not explicitly state the heatmap summary percentages and details. Start each description with \textbf{'I see'} and end with a random suitable rerouting phrase of your choice. Replace \textbf{'the image'} anywhere in your description with \textbf{'my view'}"}
\end{tcolorbox}

\subsection{Explainability Model}
The robot is equipped with a modular explainability model implemented as four ROS2 nodes, 1) Camera Node, 2) BLIP Node, 3) Heatmap Node, and 4) LLM node which will be explained in the experimental section. Each node is responsible for a distinct function. These nodes communicate through ROS topics, enabling scalable and seamless integration with existing navigation systems. This node presents information in a concise, human-understandable format, enhancing explainability in dynamic environments. The camera captures a single image on request, the LLM node must be initialized first, followed by the Heatmap and BLIP nodes.

\subsubsection{Explainability Model Formulation}
To formally define our explainability model, let \( X \) represent the raw image input captured by the robot’s camera. The explainability function \( E \) maps the visual input, the heatmap analysis, and the language model output to a structured explanation by:

\begin{equation}
E: (X, H, L) \rightarrow \mathcal{T}
\end{equation}

\noindent where, \( X \in \mathbb{R}^{m \times n \times 3} \) is the image captured at resolution \( m \times n \),  \( H = g(X) \) is the heatmap function that highlights the salient regions,  \( L = f(X, H) \) represents the captioning output of the language model and \( \mathcal{T} \) is the final textual explanation produced. The heatmap generation function \( g(X) \) is given by Grad-CAM activation \( A_c \) as

\begin{equation}
H_{i,j} = ReLU\left( \sum_k \alpha_k A_c^{i,j} \right)
\end{equation}

\noindent where, \( \alpha_k \) is the weight for the feature map \( k \), \( A_c^{i,j} \) represents the activation at the spatial location \( (i,j) \), and \( ReLU(\cdot) \) ensures positive activation contributions. The final natural language explanation \( \mathcal{T} \) is derived using

\begin{equation}
\mathcal{T} = LLM(\psi (H, X))
\end{equation}

\noindent where, \( \psi(H, X) \) is the feature representation that combines the heatmap and the image context and \( LLM(\cdot) \) is a large language model (e.g. GPT-3.5 Turbo) trained for textual summarization obtained in the LLM Guidance Prompt box. The textual explanation generated by the LLM, which depends on the human context and perception input and the variables related to the robot interface, captured as uncertainty $\mathcal{U}$, which reflects subjective interpretation, clarity of the interface and variability of trust.

\begin{equation}
\varepsilon = f\left( \mathcal{T}, \mathcal{U} \right).
\label{eq:explainability_factor}
\end{equation}
User surveys will allow determining $\varepsilon$ more precisely. 

\subsubsection{Latency Optimization for Real-Time Explainability}
Latency is critical in real-time systems. The total explanation time \( T_{\text{total}} \) is defined as:

\begin{equation}
T_{\text{total}} = T_{\text{camera}} + T_{\text{BLIP}} + T_{\text{heatmap}} + T_{\text{LLM}}
\end{equation}

\noindent where, \( T_{\text{camera}} \) is the image acquisition time, \( T_{\text{BLIP}} \) is the processing time in the vision language, \( T_{\text{heatmap}} \) is the heatmap generation time, and \( T_{\text{LLM}} \) is the time required for the large language model to generate an explanation. Since LLM processing is performed remotely, LLM request latency \( T_{\text{LLM}} \) can be modeled as

\begin{equation}
T_{\text{LLM}} = T_{\text{network}} + T_{\text{processing}}
\end{equation}

\noindent where, \( T_{\text{network}} \) represents the latency of network transmission and \( T_{\text{processing}} \) is the cloud-based inference time. To minimize \( T_{\text{total}} \), one can formulate the optimization problem as

\begin{equation}
\min_{\lambda} \sum_{i} T_i, \quad \text{s.t.} \quad T_{\text{total}} \leq T_{\text{max}}
\end{equation}

\noindent where, \( \lambda \) represents hyperparameters tuning latency trade-offs and \( T_{\text{max}} \) is the maximum allowable latency for real-time operation.

Empirical analysis showed that latency is inversely correlated with compute power \( C \):

\begin{equation}
T_{\text{total}} \propto \frac{1}{C}
\end{equation}

\noindent where increasing computing power reduces processing time. However, this research is a proof of concept and further architecture optimization, such as offloading to the edge or distributed computing, can be performed for real-world applications. 

\section{EXPERIMENTS}
To validate the effectiveness of our explainability module, we conducted structured experiments using a mobile robot running ROS 1 Noetic on a Raspberry Pi 4B with a built-in camera. The system was tested in both manual and autonomous navigation modes, with and without the explainability module active.

\subsection{Experimental Setup}
The explainability module, originally developed in ROS 2 Humble, was adapted to ROS 2 Foxy and deployed on a separate system for compatibility with the MYAGV robot. Communicated independently while generating real-time explanations. The robot was equipped with a speaker and display to provide multimodal feedback. The images were captured every 5 seconds and processed by the Camera, BLIP, Heatmap, and LLM nodes. Explanations were visualized as heatmap overlays and spoken aloud to enhance interpretability.

\subsection{Navigation and Testing Conditions}
The experiments were carried out under four scenarios:
\begin{itemize}
    \item \textbf{Manual Navigation:} As Test 1 - With and without explainability.
    \item \textbf{Autonomous Navigation:} As Test 2 - With and without explainability.
\end{itemize}
\noindent For each test, we recorded navigation metrics and explanation output, allowing us to isolate the impact of explainability as represented in Table~\ref{tab:navigation_metrics}.

\begin{table}[h!]
\centering
\begin{tabular}{|l|c|c|}
\hline
\textbf{Metric} & \textbf{WoE} & \textbf{WE} \\
\hline
\multicolumn{3}{|c|}{\textbf{Test 1: Manual Navigation}} \\
\hline
Total Trajectory (m) & 5.76 & 5.76 \\
Total Time (s) & 23.5 & 22.1 \\
Social Conflicts Detected & -- & 2 \\
Sudden Stops & 19 & 15 \\
\hline
\multicolumn{3}{|c|}{\textbf{Test 2: Autonomous Navigation}} \\
\hline
Total Trajectory (m) & 5.83 & 5.78 \\
Total Time (s) & 25.3 & 22.6 \\
Social Conflicts Detected & -- & 3 \\
Sudden Stops & 21 & 18 \\
\hline
\end{tabular}
\caption{Comparison of navigation performance under two conditions: WoE = Without Explainability, WE = With Explainability. Tests were conducted over a 14-meter delivery task (average of 4 runs in hallway and maker-space). Metrics include total trajectory length, time taken, number of social conflict detections, and sudden stops.}
\label{tab:navigation_metrics}
\end{table}

\subsection{User Survey}
During each test, the participants observed the robot and completed a post-run survey assessing trust, clarity, and transparency. These responses were used to calculate a normalized explainability factor \( \varepsilon \in [0, 1] \), with \( \varepsilon = 0 \) for non-explaining runs. We collect responses from $30$ participants, including students and faculty.

\section{ANALYSIS}
We analyze the AMR performance metrics along with \( \varepsilon \) to assess how explainability influenced navigation behavior. This included latency, stability, and confusion matrix evaluations comparing system output with human expectations. Table~\ref{tab:survey_results} summarizes the responses to Test 2, showing a significant increase in user trust and understanding when explanations were provided. We computed the overall preference score using the following.

\begin{equation}
\text{PS} = \frac{U + 0.5N}{T} \times 100,
\label{eq:preference_score}
\end{equation}

\noindent where, \( U = 22 \) (users who prefer explanations), \( N = 6 \) (neutral responses), and \( T = 30 \) (total participants), resulting in a PS of $76.7\%$. Figures~\ref{fig:survey1} and~\ref{fig:survey2} (Test 1 and Test 2, respectively) highlight a notable improvement in trust ($+16. 7\%$), understanding ($+23. 3\%$) and overall preference (from $50\%$ to $76.7\%$) when explanations were enabled.

\begin{table}[h!]
\centering
\begin{tabular}{|p{3.5cm}|c|c|c|}
\hline
\textbf{Question} & \textbf{Yes (\%)} & \textbf{Neutral (\%)} & \textbf{No (\%)} \\
\hline
The robot’s explanations helped me 
understand its decisions & 73.3\% & 20\% & 6.7\% \\
\hline
The information provided by the 
robot was clear and useful & 76.7\% & 16.7\% & 6.6\% \\
\hline
The robot’s explanations increased
my trust in it & 66.7\% & 16.7\% & 16.7\% \\
\hline
I felt more in control when

explanations were given & 70\% & 26.7\% & 3.3\% \\
\hline
\end{tabular}
\caption{Survey results measuring user trust}
\label{tab:survey_results}
\end{table}

\subsection{Explanation Latency Analysis}
The latency from module initialization to LLM summary display was measured in 88 samples, ranging from 5.986 to 50.688~s, with an average of approximately 20~s. Manual triggering significantly reduced high-latency occurrences compared to fixed 25-second intervals. The system, running on a Raspberry Pi 4 Model B (quad-core Cortex-A72, 4GB RAM), demonstrates that hardware limitations contribute to processing delays, suggesting that future upgrades may yield sub-5s latency. Higher latency directly impacts the explainability factor \( \varepsilon \), as delayed explanations reduce user trust, perceived system responsiveness, and transparency. In real-time navigation, if the robot’s explanation arrives too late relative to its decision, users may find the behavior confusing or untrustworthy. Thus, minimizing latency is critical to maintaining high \( \varepsilon \) scores in user evaluation. 

\begin{figure}[h!]
    \centering
    \includegraphics[width=1\linewidth]{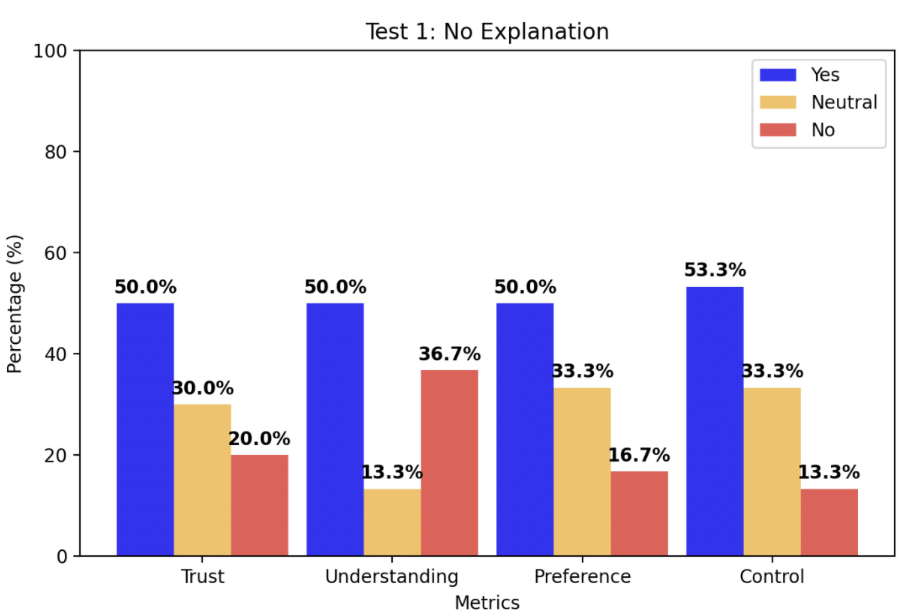}
    \caption{Test 1: User survey results.}
    \label{fig:survey1}
\end{figure}

\begin{figure}[h]
    \centering
    \includegraphics[width=1\linewidth]{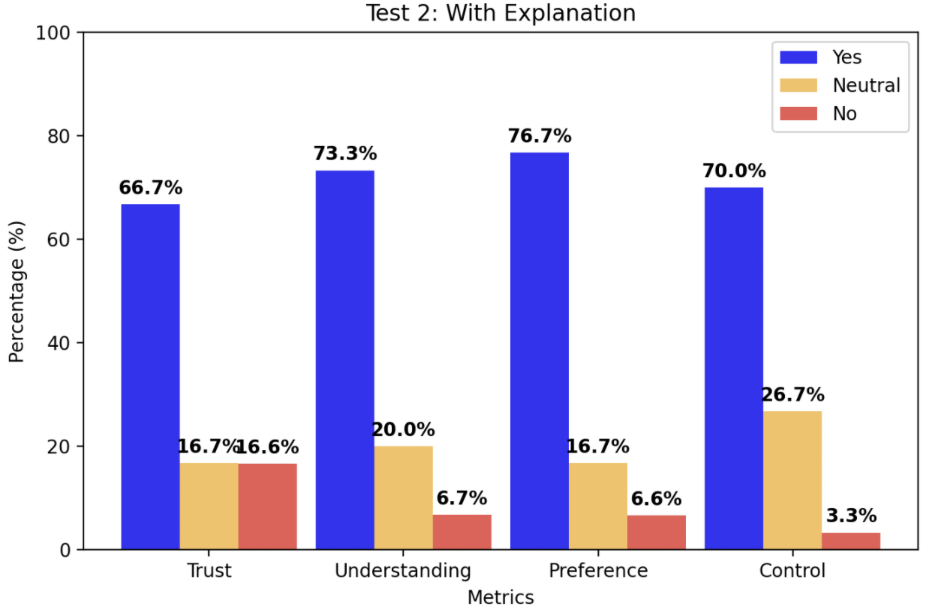}
    \caption{Test 2: User survey results.}
    \label{fig:survey2}
\end{figure}

\subsection{Response Consistency via Confusion Matrix}
We evaluated the precision of the explanation by comparing the model output with ground truth labels. Table~\ref{tab:performance_metrics} shows the confusion matrix with 196 evaluated images ($TP = 82$, $FN = 15$, $FP = 20$, $TN = 79$). Performance metrics were computed as follows.

\begin{equation}
\text{Accuracy} = \frac{TP + TN}{TP + FP + FN + TN} = 82.14\%,
\label{eq:accuracy}
\end{equation}

\begin{table}[h!]
\centering
\begin{tabular}{|c|c|c|}
\hline
\textbf{} & \textbf{Predicted Positive} & \textbf{Predicted Negative} \\
\hline
\textbf{Actual Positive} & TP: 82 & FN: 15 \\
\textbf{Actual Negative} & FP: 20 & TN: 79 \\
\hline
\end{tabular}
\caption{Confusion matrix showing performance of the explainability module. True Positive (TP), False Positive (FP), False Negative (FN), True Negative (TN).}
\label{tab:performance_metrics}
\end{table}


\section{CONCLUSIONS} 
\label{sec:conclusion}
This study demonstrates that the integration of social context awareness using visual and language models as an explainability module into mobile robot navigation significantly improves performance and social acceptance in collaborative environments between humans and robots. The survey results and experimental evaluations confirm that real-time explanations improve trust, interpretability, and transparency by aligning robot behavior with human expectations and reducing uncertainty. The high accuracy of the system and the F1 score further validate its effectiveness in addressing the black-box limitations of AI. Although latency remains a challenge, results show that optimized explanation delivery contributes to more predictable and user-aligned robotic actions.

\section*{Acknowledgments}
The authors thank Prof. Katsuo Kurabayashi and Dr. Rui Li for their invaluable feedback. We also appreciate the support and encouragement of our colleagues in the Department of Mechanical and Aerospace Engineering, New York University.

\bibliographystyle{unsrt}
\bibliography{references}

\appendix
\subsection*{Explainability Architecture in ROS}
\subsubsection*{Camera Node}
The Camera Node captures images on demand, saving and publishing them to \textit{/camera/imageRaw} for processing. This ensures optimized computational resources while providing the necessary visual input for the explainability module.
\subsubsection*{BLIP Node}
The BLIP Node processes images using Bootstrapped Language Image Pretraining (BLIP) to generate a contextual caption describing the image content. It subscribes to the \textit{/camera/imageRaw} topic to retrieve images and runs the BLIP model using the Hugging Face API due to its high computational requirements. The generated caption is published on the \textit{/blip/caption} topic, where other nodes can access it. This step bridges the gap between raw visual input and human-readable descriptions. Algorithm~\ref{alg:BlipNode} shows the pseudocode of the node working process.
\subsubsection*{Heatmap Node}
The Heatmap Node visualizes the most relevant regions of the image that influenced the captioning of the BLIP model. It applies Grad-CAM (Gradient-weighted Class Activation Mapping) with a ResNet model to highlight image areas that contribute the most to the BLIP output. In addition to generating the heatmap overlay, the node calculates the percentage of the image that the model focuses on and publishes this as a concise summary of the \textit{/heatmap/summary} topic. This provides quantitative insights into the influence of different image regions, enhancing transparency in decision-making. Algorithm~\ref{alg:HeatmapNode} shows the pseudocode of the node’s working process.
\subsubsection*{LLM Node}
The LLM Node generates a natural language explanation of the robot’s surroundings and decision-making rationale. It subscribes to both /blip/caption and /heatmap/summary, merging these outputs to form a coherent, structured response. A guiding prompt is used to ensure that the explanation follows a consistent and understandable format. Due to the high computational demands of GPT-3.5 Turbo, the processing is offloaded to the Azure OpenAI API, ensuring efficient real-time response generation. Algorithm~\ref{alg:LLMNode} shows the pseudocode of the node’s working process.
\begin{algorithm}[h!]
	\caption{BLIP Node Image Captioning}
	\label{alg:BlipNode}
	\nl Initialize the BLIP Node\;
	\nl Subscribe to topic \texttt{'camera/image\_raw'}\;
	\While{new image received}{
		\nl Extract features from image\;
		\nl Generate caption using  VLM\;
		\nl Publish caption to topic \texttt{'blip/caption'}\;
		\nl \If{publish successful}{Log success}
		\Else{Log failure}
	}
\end{algorithm}
\vspace{-3mm}
\begin{algorithm}[h!]
	\caption{Heatmap Node Processing}
	\label{alg:HeatmapNode}
	\nl Initialize the Heatmap Node\;
	\nl Subscribe to topic \texttt{'camera/image\_raw'}\;
	\While{new image received}{
		\nl Process image to generate heatmap overlay\;
		\nl Save heatmap image\;
		\nl Publish heatmap summary to topic \texttt{'heatmap/summary'}\;
		\nl \If{publish successful}{Log success}
		\Else{Log failure}	}
\end{algorithm}
\vspace{-3mm}
\begin{algorithm}[h!]
	\caption{LLM Node Explanation Generation}
	\label{alg:LLMNode}
	\nl Initialize the LLM Node\;
	\nl Subscribe to topics \texttt{'blip/caption'} and \texttt{'heatmap/summary'}\;
	\While{caption and heatmap summary received}{
		\nl Generate textual explanation using LLM\;
		\nl Synthesize speech output from the generated explanation\;
		\nl Display and save explanation with heatmap overlay\;
		\nl Save image and heatmap with timestamp for validation\;
		\nl \If{processing successful}{Log success}
		\Else{Log failure}	}
\end{algorithm}
\end{document}